# Data Sets: Word Embeddings Learned from Tweets and General Data

Quanzhi Li, Sameena Shah, Xiaomo Liu, Armineh Nourbakhsh

Research and Development
Thomson Reuters
3 Times Square, NYC, NY 10036
{quanzhi.li, sameena.shah, xiaomo.liu, armineh.nourbakhsh}@thomsonreuters.com

**Abstract**

A word embedding is a low-dimensional, dense and real-valued vector representation of a word. Word embeddings have been used in many NLP tasks. They are usually generated from a large text corpus. The embedding of a word captures both its syntactic and semantic aspects. Tweets are short, noisy and have unique lexical and semantic features that are different from other types of text. Therefore, it is necessary to have word embeddings learned specifically from tweets. In this paper, we present ten word embedding data sets. In addition to the data sets learned from just tweet data, we also built embedding sets from the general data and the combination of tweets with the general data. The general data consist of news articles, Wikipedia data and other web data. These ten embedding models were learned from about 400 million tweets and 7 billion words from the general text. In this paper, we also present two experiments demonstrating how to use the data sets in some NLP tasks, such as tweet sentiment analysis and tweet topic classification tasks.

## 1. Introduction

Distributed representations of words are also called word embeddings or word vectors. They help learning algorithms achieve better performance in natural language processing (NLP) related tasks by grouping similar words together, and have been used in lots of NLP applications, such as sentiment analysis (Socher et al. 2013; Mass et al. 2014; Tang et al. 2014, Li et al. 2016c), text classification (Matt 2015; Li et al. 2016a), and recommendation (Li et al. 2016b).

Traditional bag-of-words and bag-of-n-grams hardly capture the semantics of words, or the distances between words. This means that words "walk," "run" and "eat" are equally distant in spite of the fact that "walk" should be closer to "run" than "eat" semantically. Based on word embeddings, "walk" and "run" will be very close to each other. In this study, the word embedding representation model is computed using a neural network, and generated from a large corpus – billions of words – without supervision. The learned vectors explicitly encode many linguistic regularities and patterns, and many of these patterns can be represented as linear translations. For example, the result of a vector calculation v("Beijing") – v("China") + v("France") is closer to v("Paris") than any other word vector.

Tweets are noisy, short and have different features from other types of text. Because of the advantages of applying word embeddings in NLP tasks, and the uniqueness of tweet text, we think there is a need to have word embeddings learned specifically from tweets (TweetData). Tweets cover various topics, and spam is prevalent in tweet corpora. In addition to spam, some tweets are semantically incoherent or nonsensical, contain nothing but profanity, and are focused on daily chitchat or advertisement, etc. We consider both spam and such tweets with no substantial content as "spam" (or "noise"). Vector models built on these types of tweets will bring lots of noise to some applications. To build vector models without using spam tweets, we use a spam filter to remove the spam tweets. But on the other hand, some applications may need embeddings generated from all tweets, including the regular tweets and spam tweets. To build accurate embedding models, another question to address is whether to include phrases or not. Here a phrase means a multi-word term. An embedding model (or data set) for just words is much smaller than that for both words and phrases, and it will be more efficient for some applications that only need embeddings for words. In this study, we use a data-driven approach to identify phrases. To accommodate various applications and use cases, we generated four embedding sets using just TweetData, which are the four combinations of with/without spam tweets and with/without phrases

In some cases, such as the applications that need to work with both tweets and regular text, we may need word embeddings that are trained on both TweetData and general text data (GeneralData). Therefore, we also built four em-bedding data sets that were learned from the training data that combine TweetData and GeneralData. These four embedding sets are also from the combinations of with/without spam tweets and including phrases or not. In addition to these eight data sets, two data sets (with and without phrases) that are built from just GeneralData are also provided. Overall, ten word embedding data sets are presented in this paper.

The major contributions of this paper are:
- We publish ten word embedding data sets learned from about 400 million tweets and 7 billion words from general data. They can be used in tasks involving social media data, especially tweets, and other types of textual data. Users can choose different embedding sets based on their use cases; they can also easily try all of them to see which one provides the best performance for their application.
- We also use two experiments to demonstrate how to use these embeddings in practical applications. The second experiment also shows the performance difference of the tweet topic classification task on several data sets.

In the following sections, we first describe the technologies used for generating the word embedding data sets, then we present the data collections and preprocessing steps for both TweetData and GeneralData, followed by the descriptions of the ten data sets, and finally we present the two experiments.

## 2. Technologies Used for Building Word Embedding Data Sets

In this section we describe three technologies used in building our word vector models: distributed word representation, phrase identification and tweet spam filtering

### 2.1 Distributed Word Representation

A distributed language representation $X$ consists of an embedding for every vocabulary word in space $S$ with dimension $D$, where $D$ is the dimension of the latent representation space. The embeddings are learned to optimize an objective function defined on the original text, such as likelihood for word occurrences. Word embedding models have been researched in previous studies (Collobert et al. 2011; Mikolov et al. 2013b; Socher et al. 2014). Collobert et al. (2011) introduce the *C&W* model to learn word embeddings based on the syntactic contexts of words. Another implementation is the word2vec model from (Mikolov et al. 2013a, 2013b). This model has two training options, Continuous Bag of Words and the Skip-gram model. The Skip-gram model is an efficient method for learning high-quality distributed vector representations that capture a large number of precise syntactic and semantic word relationships. Based on previous studies and the experiments we conducted in other tasks, the Skip-gram model produces better results, and here we briefly introduce it.

The training objective of the Skip-gram model is to find word representations that are useful for predicting the surrounding words in a sentence or a document. Given a sequence of training words $W_1, W_2, W_3, \ldots, W_N$, the Skip-gram model aims to maximize the average log probability.

$$\frac{1}{N}\sum_{n=1}^{N} \sum_{-m \leq i \leq m, i \neq 0} \log p(W_{n+i} \mid W_n)$$

where $m$ is the size of the training context. A larger $m$ will result in more training data and can lead to a higher accuracy, at the expense of the training time.

Generating word embeddings from text corpus is an unsupervised process. To get high quality embedding vectors, a large amount of training data is necessary. After training, each word, including all hashtags in the case of tweet text, is represented by a real-valued vector. Usually the dimension size ranges from tens to hundreds.

### 2.2 Phrase Identification

Phrases usually convey more specific meanings than single-term words. In many phrases, each has a meaning that is not a simple composition of the meanings of its individual words. Therefore, it is important to also learn vector representation for phrases, which are very useful in many applications. To identify phrases from TweetData and GeneralData, we use the approach described in (Mikolov et al. 2013b). We first find words that appear frequently together, and infrequently in other contexts. For example, "New York City" is replaced by a unique token in the training data, while a bigram "I am" will remain untouched. The good thing about this approach is that we can form many reasonable phrases without greatly increasing the vocabulary size. In contrast, if we train the Skip-gram model using all n-grams, it would be very memory intensive. To identify phrases, a simple data-driven approach is used, where phrases are formed based on the unigram and bigram counts, using this scoring function:

$$Score(w_i, w_j) = \frac{C(w_i, w_j) - \mu}{C(w_i) * C(w_j)}$$

Where $C(w_i, w_j)$ is the frequency of word $w_i$ and $w_j$ appearing together. $\mu$ is a discounting coefficient to prevent too many phrases consisting of infrequent words to be generated. The bigrams with score above the chosen threshold are then used as phrases. Then the process is repeated 2-4 passes over the training data with decreasing threshold value, so we can identify longer phrases consisting of several words. The maximum length of a phrase is limited to 4 words in our data sets. Other parameters are set as the default values used in (Mikolov et al. 2013b), and the code is available at: (https://code.google.com/p/word2vec).

## 2.3 Tweet Spam Filter

We use the tweet "spam" filtering algorithm described in (Liu et al. 2016) to identify spam tweets (or noise tweets). It is more than just getting rid of standard spam. There are a lot of tweets that carry little semantic substance, such as profanity, chit-chat, advertisements, etc. The goal of this spam filter is to mark all of these categories as spam, and only preserve the informative tweets that contain information of some interest. The spam filtering algorithm is a hybrid approach that combines both rule-based and learning-based methods. Inspired by studies of Yardi et al. (2009) and Song et al. (2011), this approach uses features of follower-to-friend relationship, tweet publication frequency, and other indicators to detect standard spam. The profanity and advertisement can be largely removed using keyword lists. The chit-chat is identified by a language model trained with conversational SMS messages. These steps can help us chop off a significant amount of unsubstantial tweets. We are able to obtain a set of informative tweets after these filtering steps. We basically use the following two steps:

*Spam & Advertisement Filtering*: The metadata of each tweet contains detailed information about the author's profile and the tweet's content. Given that Twitter has already prevented spam with harmful URLs (Thomas et al. 2011), we only concentrate on signals from user profiles and tweet content. Since our spam and advertisement filtering algorithms share the same strategy, we combine their introduction here. We followed the ideas from DataSift Inc. (http://dev.datasift.com/docs/platform/csdl) to design a set of practical filtering rules (see examples below). To avoid hard cut-offs, each filtering rule thresholds a tweet into three levels: ``noisy", ``suspicious" and ``normal". At least one ``noisy" or two-plus ``suspicious" rules can mark a tweet as ``noisy".

*Conversation Filtering*: Following the approach in (Balasubramanyan and Kolcz 2013), a topic model was trained on two online conversation corpora with 20,584 messages. Chat topics were manually identified. When a new tweet arrives, its posterior topic distribution is inferred. If the probability mass concentrates on the chat topics is above a threshold, this tweet is recognized as a chat tweet. In addition, a few heuristics are applied to enhance the chit-chat detection such as if a tweet contains multiple first & second personal pronouns, emoticons and emojis.

## 3. Word Embedding Data Sets

In this section, we first introduce the two training data collections, TweetData and GeneralData, used for generating the word embeddings, and the basic steps to preprocess them. Then we introduce our ten word embedding data sets (models), listed in Table 1. Their names are self-explanatory. Here *with/without spam* means whether or not the spam tweets are included in the training data. *Word+Phrases* means this embedding data set contains both words and phrases. *TweetDataWithoutSpam+GeneralData* means we use both TweetData (without spam) and GeneralData for training this embedding model.

These ten data sets cover all the eight embedding sets involving TweetData, which are the combinations of using spam tweets or not, including phrases or not, and integrating with GeneralData or not. In addition to these eight data sets, two embedding sets (Dataset 5 and 6) using only GeneralData are also generated. After the training data is preprocessed, the word2vec's Skip-gram model (Mikolov et al. 2013b) is used to learn the word vector models. Each data set will be explained in Section 3.3.

*Table 1. The 10 word embedding data sets*

| Dataset No | Name |
|---|---|
| 1 | TweetDataWithoutSpam_Word |
| 2 | TweetDataWithoutSpam_Word+Phrase |
| 3 | TweetDataWithSpam_Word |
| 4 | TweetDataWithSpam_Word+Phrase |
| 5 | GeneralData_Word |
| 6 | GeneralData_Word+Phrase |
| 7 | TweetDataWithoutSpam+GeneralData_Word |
| 8 | TweetDataWithoutSpam+GeneralData_Word+Phrase |
| 9 | TweetDataWithSpam+GeneralData_Word |
| 10 | TweetDataWithSpam+GeneralData_Word+Phrase |

## 3.1 TweetData and the Preprocessing Steps

The tweets used for building our embedding models date from October 2014 to October 2016. They were acquired through Twitter's public 1% streaming API and the Decahose data (10% of Twitter's streaming data) obtained from Twitter. We randomly selected tweets from this period, to make them more representative. Only English tweets are used to build word embeddings. Totally, there are 390 million English tweets. Based on our spam filter, there are about 198 million non-spam tweets and 192 million spam tweets. The training data size will affect the quality of word vectors. Our experiments based on tweet topic classification and words similarity computation tasks show that after the number of training tweets exceeds 100 million, the performance changes little. But from 5,000 to 1 million tweets, the boost is very significant. Detailed information for each embedding data set is presented in Table 2, 3 & 4.

Each tweet used in the training process is preprocessed as below, and the resulting data is then processed by each of the eight model building processes involving TweetData. The preprocess steps are as follows:
- Tweet text is converted to lowercase.
- All URLs are removed. Most URLs are short URLs and located at the end of a tweet.
- All mentions are removed. This includes the mentions appearing in a regular tweet and the user handles at the beginning of a retweet, e.g. ``RT: @realTrump''.
- Dates and years, such as ``2017'', are converted to two symbols representing date and year, respectively.
- All ratios, such as ``3/23'', are replaced by a special symbol.
- Integers and decimals are normalized to two special symbols.
- All special characters, except hashtags symbol #, are removed.

These preprocessing steps are necessary, since most of the tokens removed or normalized are not useful, such as mentions and URLs. Keeping them will increase the vector space size and computing cost. Stop words are not removed, since they provide important context in which other words are used. Stemming is not applied to words, since some applications of these data sets may want to use the original forms of the words. For applications that require the same embedding for all the variations of a word, they can combine their embeddings to generate a unified one, e.g. using the average of all their embeddings.

### 3.2 GeneralData and the Preprocessing Steps

A pre-built word embedding model is provided by Mikolov et al. (2013b), and it was trained on part of Google News dataset (https://code.google.com/p/word2vec/). This model contains 300-dimensional vectors for about 3 million unique words and phrases. The phrases were obtained using the same data-driven approach described in Section 2.2. Although this model was trained on a large set of data, it has some limitations: (1) it was trained on only news articles, the words and phrases in this mode are case-sensitive; and (2) the texts were not cleaned before they were fed to the training algorithm. Therefore, there are lots of junk tokens in this model and they may affect the performance of applications using it. We discuss this problem in our second experiment. Another reason for constructing our own GeneralDatacollection is that we wanted to combine TweetData and GeneralData together as training data to learn word embeddings, which will be more appropriate for use cases that deal with both tweet data and other general text data. The news articles used for training the Google News word2vec vector model are not available to public. To build our own GeneralData collection and make it more representative, but not biased toward only one type of text, such as news articles, we collected data from five different sources, which have different types of text. The five data sources are:
- Reuters' news articles: 10 billion bytes of news data from year 2007 to 2015 were collected from Reuters' news archive.
- First billion characters from Wikipedia (http://mattmahoney.net/dc/enwik9.zip)
- The latest Wikipedia dump. There are about 3 billion words. (http://dumps.wikimedia.org/enwiki/latest/enwiki-latest-pages-articles.xml.bz2).
- The UMBC web-base corpus. There are around 3 billion words. (http://ebiquity.umbc.edu/blogger/2013/05/01/umbc-webbase-corpus-of-3b-english-words/)
- The "One Billion Word Language Modeling Benchmark" data set. It contains almost 1 billion words. (http://www.statmt.org/lm-benchmark/1-billion-word-language-modeling-benchmark-r13output.tar.gz)

The preprocessing steps are as following:
- All data is converted to plain text and lowercased.
- All URLs are removed from the text.
- Dates and years are converted to two symbols representing date and year, respectively.
- All ratios are replaced by a special symbol.
- Integers and decimals are normalized to two special symbols.
- All special characters are removed.

These preprocesses make sure that the general data is compatible with the tweet data, since we need to combine these two types of data for learning four of the ten embedding models. Changing all characters to lowercase will incorrectly merge some terms together, such as an entity name and a regular word which happen to have the same spelling, but this only affects a very small portion of the terms. Without case folding, most words will have two entries with different embeddings in the vector model, which will affect the vector quality, and the term space will increase greatly. However, it is possible that in some special use cases, such as identifying named entities from text, keeping the character cases may be a better strategy.

### 3.3 Word Embedding Data Sets and Metadata

#### 3.3.1 Dataset1: TweetDataWithoutSpam_Word

For this vector model, the training tweets do not contain spam tweets, and the GeneralData collection is not used. We only consider single-term words and no multi-word phrases are included. The basic statistics for the training data set and the metadata of the embedding vector model are shown in Table 2. About 200 million tweets are used for building this model. Totally, 2.8 billion words are pro-cessed. With a term frequency threshold of 5 (tokens less than 5 occurrences in the training data set are discarded), the total number of unique tokens (hashtags and words) in this model is 1.9 million.

*Table 2. Metadata of embedding data sets using only TweetData*

| Metadata | Dataset | | | |
|---|---|---|---|---|
| | Dataset1: TweetDataWithoutSpam_Word | Dataset2: TweetDataWitoutSpam_Word+Phrase | Dataset3: TweetDataWithSpam_Word | Dataset4: TweetDataWithSpam_Word+Phrase |
| Number of Tweets | 198 million | 198 million | 390 million | 390 million |
| Number of words in training data | 2.8 billion | 2.8 billion | 4.9 billion | 4.9 billion |
| Number of unique words or (words+phrases) in the trained embedding model | 1.9 million | 2.9 million | 2.7 million | 4 million |
| Vector dimension size | 300 | 300 | 300 | 300 |
| Word and phrase frequency threshold | 5 | 10 | 5 | 10 |
| Learning context windows size | 8 | 8 | 8 | |

*Table 3. Metadata of embedding data sets using only GeneralData*

| Metadata | Dataset | |
|---|---|---|
| | Dataset5: GeneralData_Word | Dataset6: GeneralData_Word+Phrase |
| Number of words in training data | 6.7 billion | 6.7 billion |
| Number of unique words or (words+phrases) in the trained embedding model | 1.4 million | 3.1 million |
| Vector dimension size | 300 | 300 |
| Word and phrase frequency threshold | 5 | 8 |
| Learning context windows size | 8 | 8 |

*Table 4. Metadata of embedding data sets using both TweetData and GeneralData*

| Metadata | Dataset | | | |
|---|---|---|---|---|
| | Dataset7: TweetDataWithoutSpam+GeneralData_Word | Dataset8: TweetDataWithoutSpam+GeneralData_Word+Phrase | Dataset9: TweetDataWithSpam+GeneralData_Word | Dataset10: TweetDataWithSpam+GeneralData_Word+Phrase |
| Number of Tweets | 198 million | 198 million | 390 million | 390 million |
| Number of words from GeneralData | 6.7 billion | 6.7 billion | 6.7 billion | 6.7 billion |
| Number of words in the whole training data | 9.5 billion | 9.5 billion | 11.6 billion | 11.6 billion |
| Number of unique words or (words+phrases) in the trained embedding model | 1.7 million | 3.7 million | 2.2 million | 4.4 million |
| Vector dimension size | 300 | 300 | 300 | 300 |
| Word and phrase frequency threshold | 10 | 15 | 10 | 15 |
| Learning context windows size | 8 | 8 | 8 | |

The word embedding dimension size is set to 300. The dimension size will affect the quality of a vector. Generally, larger dimension size will give better quality. We conducted experiments to see how the performance changes with different sizes of word vector, for tweet topic classification and word similarity computation tasks. Our experiments show that after the vector size reaches 300, the performance does not change significantly when the size increases. But from sizes 10 to 100, the performance improvement is very noticeable. The learning window size (sliding context window for a word) is set to 8. As mentioned before, a larger window size will result in more training data and can lead to a higher accuracy, at the expense of training time. Usually, a windows size of 5 to 8 is a good choice.

### 3.3.2 Dataset2: TweetDataWithoutSpam_Word+Phrase

As explained earlier in this paper, phrases are needed in some NLP related tasks, and embeddings for phrases may help those applications. Section 2.2 describes an approach we use to identify phrases from training data. The same training data set from previous section is used for building this embedding model. The only difference is that, for this model, we first identify and mark phrases from the training data, and then the tweets with detected phrases are fed into the training process to generate embeddings for both words and phrases.

Table 2 presents the related metadata of this model. There are 2.9 million unique words and phrases in this vector model. The frequency threshold for words and phrases is set to 10 in this model, greater than 5 used in Dataset1. This setting reduces the size of the model, since it has both words and phrases.

In the final embedding file, the words in a phrase are connected by "_" instead of a space. For example, "new york" will be "new_york." When looking up a phrase in the embedding model, users need to first convert the space between the phrase's words to "_".

### 3.3.3 Dataset3: TweetDataWithSpam_Word

Some applications may need embeddings generated from all type of tweets, including spam tweets, and our spam filter may not be appropriate for some applications. Therefore, we also provide embedding data sets that are built from all the tweets in the TweetData collection. Related metadata for this model can be found from Table 2. A total of 390 million tweets are used in this model. The number of words in the training data set is 4.9 billion, and the number of unique words and phrases in the final embedding model is 2.7 million.

### 3.3.4 Dataset4: TweetDataWithSpam_Word+Phrase

Similar to Dataset3, this data set is also learned from both spam and non-spam tweets, but it includes phrases. The total number of unique terms in this model is 4 million. Compared to the last model, the number of unique terms increases significantly, due to the phrases identified.

### 3.3.5 Dataset5: GeneralData_Word

This data set is learned from the GeneralData collection, which consists of text types other than tweets. As mentioned before, we provide this set in case some users need embeddings from non-tweet data, or want to do some comparison between embeddings from tweet data and non-tweet data. This data set contains only words and its related metadata are presented in Table 3.

### 3.3.6 Dataset6: GeneralData_Word+Phrase

Similar to Dataset5, this one is also learned from just GeneralData, but it includes both words and phrases. Table 3 has the metadata for this model.

### 3.3.7 Dataset7: TweetDataWithoutSpam+GeneralData_Word

Some applications or NLP tasks may need to deal with both tweet data and general-domain data, and a word embedding data set combining these two types of data may provide better performance than the one learned from just one of them. This is the motivation behind this dataset, as well as also datasets 8, 9, and 10. From Table 4, we can see that there are 1.7 millions unique words in this data set. Compared to using just TweetData as the training data, we have 6.7 billion more words from GeneralData included in the training data. With a word frequency threshold 10, the unique number of words in the final embedding data set is 1.7 million.

### 3.3.8 Dataset8: TweetDataWithoutSpam+GeneralData_Word+Phrase

The only difference between this data set and Dataset7 is that this one includes both words and phrases. Consequently, the total number of unique terms in this set is much larger, 3.7 million, than that in Dataset7, which is 1.7 million, even though this model has a larger term frequency threshold, 15. Table 4 contains the related metadata for this model.

### 3.3.9 Dataset9: TweetDataWithSpam+GeneralData_Word

This set uses both spam and non-spam tweets from TweetdData, together with GeneralData for training. It contains embeddings for only words. Related metadata are also presented in Table 4.

### 3.3.10 Dataset10: TweetDataWithSpam+GeneralData_Word+Phrase

Compared to Dataset9, this one includes both words and phrases. Table 4 shows that this data set has the most unique terms, 4.4 million.

## 3.4 How to Retrieve the Embeddings

Each published embedding data set includes a binary file, which contains the words (and phrases, for data sets including phrases) and their embeddings. It also contains a text file, which contains a list of all the words (and phrases) in this data set and their frequencies. The text file is just for reference purposes users can just use the binary embedding file without the text file.

There are several options to retrieve the embeddings from the binary model files. Here we just list some:
- Use Python's gensim package, which has the implementation of word2vec, available at: http://radimrehurek.com/gensim/.
- Use a Java implementation of word2vec, available at: http://deeplearning4j.org/word2vec.html.
- Use a C++ version of word2vec, available at: https://code.google.com/archive/p/word2vec/
- Use the Python script we provide with the published data sets.

The basic steps of looking up words or phrases from these models (data sets) are very simple. (1) Load the model into memory through one of the above methods. (2) The model will be stored as a map with words or phrases as the keys, and their embeddings (each represented as a list of 300 real numbers), as the map values. Then one can retrieve the embedding of a term by looking up the map. If a term does not exist in this data set, the lookup will return a null value. One simple solution for the non-existing terms is to set their embedding to zero.

In addition to storing the embedding model as binary file, we can also store them as plain text file, but the file size would be very big ( at least several gigabytes). If any user needs the plain text format of data, we can also provide it to them.

## 4. Experiments

We conducted two experiments: the first one uses a tweet sentiment analysis task to show how to use the word embedding data set, and the second one tests four embedding data sets on tweet a topic classification task to show their performance difference.

### 4.1 Experiment on Tweet Sentiment Analysis

In this experiment, we show how the word embeddings can be used in a real NLP task, which is tweet sentiment analysis. We are not going to compare our approach to other methods, since that is out of the scope of this paper. In this experiment, we just use one vector model, Dataset1: TweetDataWithoutSpam_Word, to demonstrate how to use it. The other embedding sets can be used in the same way.

For tweet sentiment analysis, we evaluate precision, recall, F measure and accuracy. It is a two-way classification task, i.e. we have two polarities: positive and negative.

#### 4.1.1 Sentiment Analysis Data Set

The experiment is conducted on a benchmark data set, which is from task 9 of SemEval 2014 (Sara et al. 2014). Its training set is the same as that of task 2 in SemEval 2013 (Nakov et al. 2013). The test set includes the test data from 2013 and some new added tweets. The development set is used for model development, and fine-tuning model parameters. The actual tweet texts are not provided in these data sets, due to privacy concerns. So we downloaded these tweets from Twitter's REST API for this experiment. Table 5 shows the distributions of the downloaded tweets.

*Table 5. Sentiment Analysis Dataset Statistics*

| Dataset | Positive | Negative | Total |
|---|---|---|---|
| **Train** | 2,294 | 853 | 3,147 |
| **Development** | 969 | 322 | 1,291 |
| **Test** | 1,588 | 439 | 2,027 |

#### 4.1.2 Word Embeddings for Tweet Representation

Given a tweet, we process it by the following steps:
- First, use the same steps in Section 3.1 to preprocess the tweet, and get its cleaned text.
- Second, for each word, we look up its embedding from the vector model. The result is a 300-dimension vector of real values. If a token is not contained in the model, we can either ignore that token, use a vector whose values are all 0 to represent this token, or use the average of the embeddings from words having the lowest frequency in the model. In this experiment, we just ignore the token if it does not exist in the model file. Usually they are misspelled words.
- Now, for each word of this tweet, we have a real-value vector. Because tweets have different lengths, we need to use a fixed-length vector to represent a tweet, so that we can use it in any learning algorithm or application. The following paragraph describes how we produce a tweet representation from the embeddings of its words.

**Tweet Representation:** There are different ways to obtain tweet representation from word embeddings. The most common methods use the maximum (max), minimum (min), or average (ave) of the embeddings of all words (or just the important words) in a tweet (Socher et al. 2014; Tang et al. 2014). Take the max as an example, to produce the max vector, for each of the 300 dimensions, we use the maximum value of all the word embeddings of this tweet on this dimension. In this study, we try all these three methods, and also the concatenation convolutional layer (con), which concatenates max, min and ave together. The concatenation layer is expressed as follow:

$$Z(t) = [Z_{max}(t), Z_{min}(t), Z_{ave}(t)]$$

where $Z(t)$ is the representation of tweet $t$.

For max, min and ave approaches, the dimension of a tweet vector is 300. For the con approach, the dimension size for a tweet is 900, since we concatenate three 300-dimension vectors together.

A study from (Mitchell and Lapata 2008) shows that using multiplication of embeddings can also give good performance. Users can try this approach in their own applications.

### 4.1.3 Result

In this experiment, we applied several classification algorithms to find out which one performs the best, such as LibLinear, SMO (Keerthi et al. 2011; Platt 1998), Random Forest and Logistic Regression. Their performance was comparable, with the LibLinear model (Fan et al. 2008) performing slightly better. Here we present the results from LibLinear. Table 6 shows the sentiment analysis result using the four different convolution layer approaches. We can see that ave and con perform better than the max and min approaches. This result does not mean that they will perform the same way in other use cases.

*Table 6. Tweet sentiment analysis performance using word embedding*

| Method | Precision | Recall | F measure | Accuracy |
|---|---|---|---|---|
| Max | 78.6 | 80.4 | 79.1 | 80.4 |
| Min | 79.9 | 81.7 | 80.1 | 81.7 |
| Ave | 83.1 | 84.2 | 82.6 | 84.1 |
| Con | 82.6 | 83.4 | 82.8 | 83.4 |

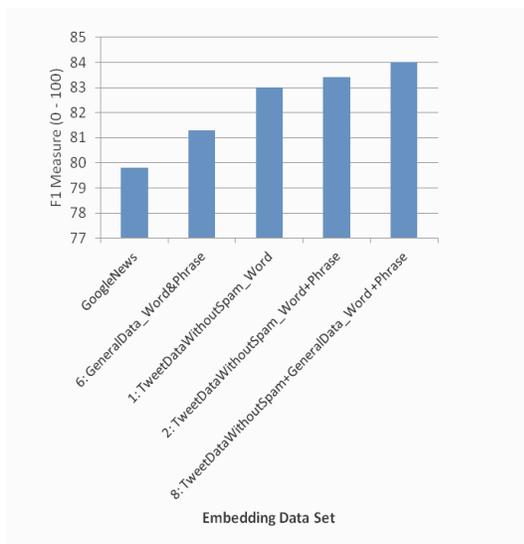

*Figure 1. Comparing embedding data sets on tweet topic classification performance*

### 4.2 Experiment on Tweet Topic Classification

This experiment is to show how some of the embedding data sets perform differently on the tweet topic classification task. We are not going to test all the ten models in this experiment. These data sets may perform differently in different applications. Users can try them in their use cases to see which one performs the best. In this experiment, we used embedding models learned from the following four data sets: Dataset1, Dataset2, Dataset6, and Dataset8. The reason for choosing these four is that from the results of these four data sets, we can perform the following comparisons: Word vs. Word+Phrase (Dataset 1 vs. Dataset 2), TweetData vs. GeneralData (Dataset2 vs. Dataset6), TweetData vs. TweetData+GeneralData (Dataset2 vs. Dataset8), and GeneralData vs. TweetData+GeneralData (Dataset6 vs. Dataset8). We use the GoogleNews word2vec data set as the baseline.

The task is to classify the test tweets into one of 11 topic categories, such as Sports, Politics, and Business. The labeled data set are from (Li et al. 2016). Totally, there are 25,964 labeled tweets, each of which belongs to one of the 11 topic categories. These tweets were split into training, validation and test sets. We use the same tweet representation method described in experiment 1, Section 4.1.2, to generate tweet embedding from its word embeddings. The tweet embeddings are the features used by classification algorithms.

We tried different classifiers, such as LibLiner and SMO, and SMO performed the best. SMO is a sequential minimal optimization algorithm for training a support vector classifier (Keerthi et al. 2011; Platt 1998). The results shown in Figure 1 are based on the SMO algorithm. From Figure 1, we can see that all of our four data sets perform better than the GoogleNews word2vec set. Dataset1, 2 and 8 perform better than Dataset6, which means, on tweet related tasks, the word embeddings learned from tweet data are better than the embedding models learned from other types of data. Dataset2 has a better result than Dataset1, which shows that including phrases in the vector model improves the classification performance. Dataset8 is the best performer; this demonstrates that combining both TweetData and GeneralData together does improve the embedding quality, and consequently, the topic classification performance.

## 5. Conclusion

Distributed word representations can benefit many NLP related tasks. This paper presents ten word embedding data sets learned from about 400 million tweets and billions of words from general textual data. These word embedding data sets can be used in Twitter related NLP tasks. Our experiments also demonstrated how to use these embeddings. Experimental results show that context is important when a classification task is at hand. For example, vectors trained on tweet data are more useful as features for a tweet classification task, than vectors trained on long form content. In addition, our experiments show that phrase detection (even though it comes at the cost of processing time) can generate more useful vectors. Lastly, noise-filtering and spam detection can be helpful pre-processing steps, especially when the task is concerned with detecting the semantic topic of tweets.